\documentclass{article}
\usepackage{microtype}
\usepackage{graphicx}
\usepackage{subfigure}
\usepackage{booktabs} 
\usepackage{wrapfig}
\usepackage{multirow}  
\usepackage{bm} \bm{}
\usepackage{caption}
\usepackage{microtype}
\usepackage{graphicx}
\usepackage{subfigure}
\usepackage{booktabs} 
\usepackage{bbm}
\usepackage{amsmath}
\usepackage{amssymb}
\usepackage{url}
\usepackage{ulem}
\usepackage{mathtools}
\usepackage{amsthm}
\usepackage{bm} \bm{}
\def\z{{\bm z}}
\def\C{{\mathcal C}}

\usepackage{stfloats}

\newcommand\ie{\textit{i.e.}}
\newcommand\eg{\textit{e.g.}}

\usepackage{hyperref}

\usepackage[accepted]{icml2025}

\usepackage{amsmath}
\usepackage{amssymb}
\usepackage{mathtools}
\usepackage{amsthm}
\usepackage{color,xcolor}
\definecolor{mygray}{gray}{.9}
\usepackage{bm} \bm{}
\def\z{{\bm z}}
\def\C{{\mathcal C}}

\def\M{{\mathcal M}}

\usepackage[capitalize,noabbrev]{cleveref}

\theoremstyle{plain}

\theoremstyle{definition}

\theoremstyle{remark}

\usepackage[textsize=tiny]{todonotes}

\icmltitlerunning{Exploring Iterative Manifold Constraint for Zero-shot Image Editing}

\begin{document}

\twocolumn[
\icmltitle{Exploring Iterative Manifold Constraint for Zero-shot Image Editing}



\icmlsetsymbol{equal}{*}

\begin{icmlauthorlist}
\icmlauthor{Maomao Li}{yyy}
\icmlauthor{Yu Li*}{comp}
\icmlauthor{Yunfei Liu}{comp}
\icmlauthor{Dong Xu*}{yyy}\\
\vspace{0.8em}
\textsuperscript{\rm 1}School of Computing and Data Science, The University of Hong Kong \quad \textsuperscript{\rm 2}International Digital Economy Academy (IDEA)
\end{icmlauthorlist}

\icmlaffiliation{yyy}{School of Computing and Data Science, The University of Hong Kong}
\icmlaffiliation{comp}{International Digital Economy Academy (IDEA)}

\icmlcorrespondingauthor{Yu Li}{liyu@idea.edu.cn}
\icmlcorrespondingauthor{Dong Xu}{dongxu@hku.hk}

\icmlkeywords{Machine Learning, ICML}
\vspace{1em}
\begin{center}
    \centering
    \captionsetup{type=figure}
    \includegraphics[width=0.95\textwidth]
{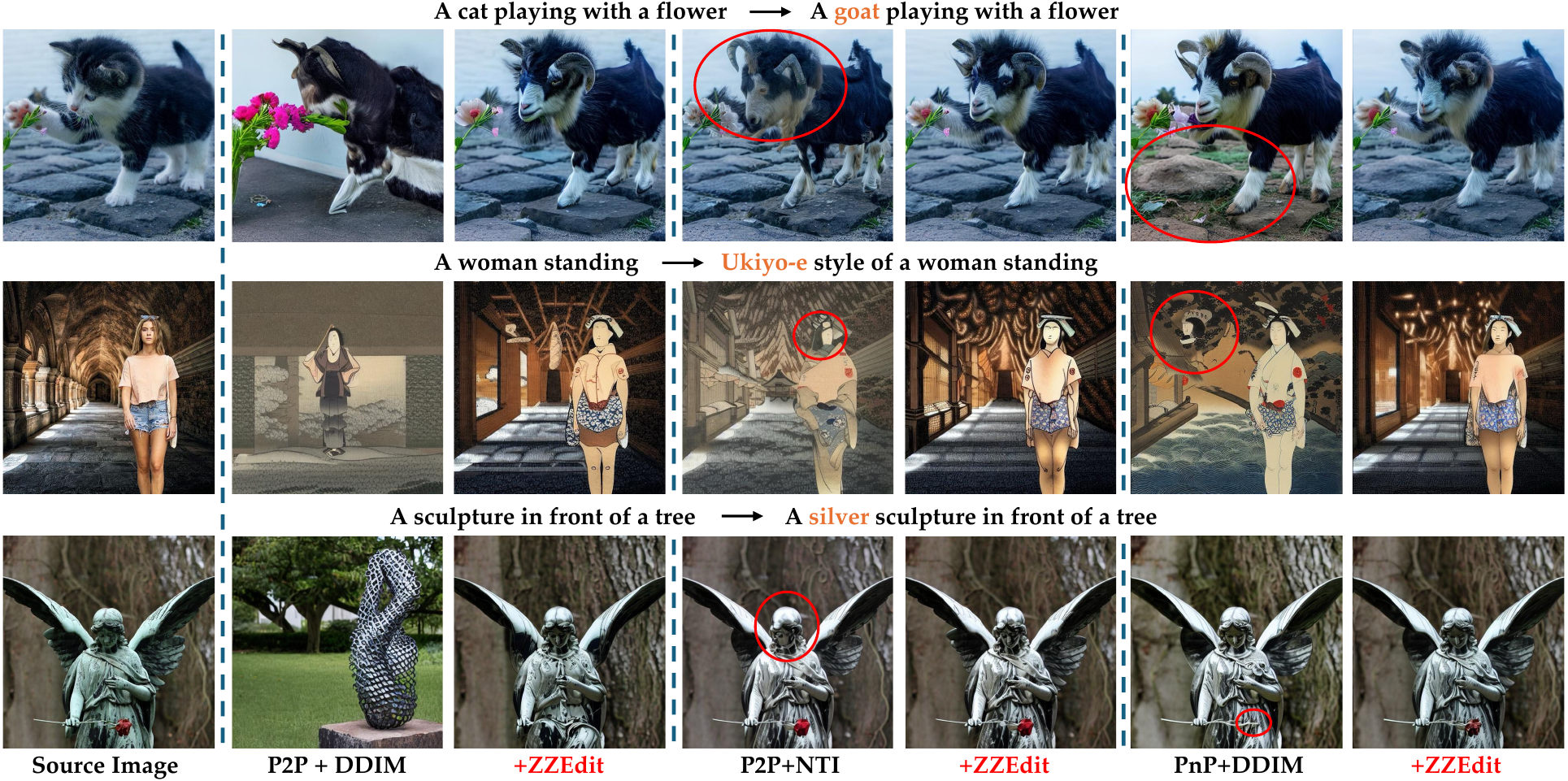}
\vspace{-0.8em}
    \captionof{figure}{\protect{We propose a novel zero-shot editing paradigm dubbed ZZEdit, which demonstrates a more subtle editability and fidelity over the commonly employed ``inversion-then-editing'' pipeline. Moreover, it seamlessly integrates with contemporary text-driven image editing methods, such as P2P~\citep{p2p} (with DDIM inversion~\citep{ddim} or Null-text inversion~\citep{null}) and PnP~\citep{pnp} (with DDIM inversion), enhancing their capabilities.} }
    \label{fig:teaser}
\end{center}%


%
%
\vskip 0.3in
]






\let\thefootnote\relax\footnotetext{*Corresponding Author}

\begin{abstract}
Editability and fidelity are two essential demands for text-driven image editing, which expects that the editing area should align with the target prompt and the rest remain unchanged separately. The current cutting-edge editing methods usually obey an "inversion-then-editing" pipeline, where the input image is inverted to an approximate Gaussian noise $\bm{z}_T$, based on which a sampling process is conducted using the target prompt. Nevertheless, we argue that it is not a good choice to use a near-Gaussian noise as a pivot for further editing since it would bring plentiful fidelity errors. We verify this by a pilot analysis, discovering that intermediate-inverted latents can achieve a better trade-off between editability and fidelity than the fully-inverted $\bm{z}_T$. Based on this, we propose a novel zero-shot editing paradigm dubbed ZZEdit, which first locates a qualified intermediate-inverted latent marked as $\bm{z}_p$ as a better editing pivot, which is sufficient-for-editing while structure-preserving. Then, a ZigZag process is designed to execute denoising and inversion alternately, which progressively inject target guidance to $\bm{z}_p$ while preserving the structure information of $p$ step. Afterwards, to achieve the same step number of inversion and denoising, we execute a pure sampling process under the target prompt. Essentially, our ZZEdit performs iterative manifold constraint between the manifold of $\M_{p}$ and $\M_{p-1}$, leading to fewer fidelity errors. Extensive experiments highlight the effectiveness of ZZEdit in diverse image editing scenarios compared with the "inversion-then-editing" pipeline.
\end{abstract}








\section{Introduction}
\label{sec:intro}
Recent years, large-scale text-guided diffusion models~\cite{imagen,ldm,dalle2,yu2022scaling,gu2022vector} have attracted growing attention in computer vision and graphics community, showing efficiency for high-quality text-to-image (T2I) synthesis. To replicate this success in text-driven image editing and enable users to manipulate input images according to their text prompt, early attempts usually take additional user-provided masks~\cite{gafni2022make,nichol2021glide,avrahami2023spatext,mokady2022self,lugmayr2022repaint} or box~\cite{li2023gligen}. Besides, \cite{zhang2023adding,qin2023unicontrol} take more conditions for fine-grained control over images e.g., depth maps, canny edges, poses, and sketches. Another line of research aims for \textit{text-only} interactive image editing~\cite{p2p,pnp,prompt,null,cao2023masactrl,ju2023direct,2022text2live,sdedit}. Since the last setting operates with minimal input conditions (i.e., only image and text) but also shows promising results for real image editing, we follow their trend in this work.


From the geometric view, image editing can described as transitions of $\M_{src} \to \M_{tgt}$, moving from a source manifold $\M^{src}_0$ to a target manifold $\M^{tgt}_0$. The current text-only image editing methods~\cite{p2p,pnp,null,prompt} usually obey the ``inversion-then-editing'' pipeline. Specifically, inversion techniques gradually add noise to the source image feature $\z_0$ (on the manifold $\M^{src}_0$) to reach an approximate Gaussian noise $\z_T$ (on the noisy manifold $\M_{T}$) as editing pivot, based on which a sampling process is carried out under the guidance of the target prompt. Here, we raise a question that \textit{is it a good choice to directly invert the input image to a near-Gaussian noise}? We believe the answer is negative from the perspective of the trade-off between \textit{editability} and \textit{fidelity}. Specifically, we conduct a pilot analysis with commonly-used DDIM inversion, and discover that intermediate-inverted latents can provide considerable editability as $\z_T$. Besides, given that DDIM inversion has accumulated errors in each step~\cite{null,prompt}, applying fully-inverted $\z_T$ for subsequent denoising would inevitably bring more reconstruction errors than intermediate-inverted ones, thus hindering the fidelity. 




Considering intermediate-inverted latents can deliver a better trade-off between \textit{editability} and \textit{fidelity}, this paper proposes a novel zero-shot editing paradigm, dubbed ZZEdit, where the insight behind is \textit{mildly strengthening guidance at a sufficient-for-editing while structure-preserving editing pivot}. Specifically, we first locate a proper intermediate-inverted latent $\z_p$ as editing pivot, which is achieved by looking up the \textit{first} step on the inversion trajectory whose response to the target prompt is greater than that to the source one. Then, we propose a ZigZag process to gentlely perform target guidance while still holding the structure information on the selected pivot $\z_p$. Concretely, our ZigZag process performs one-step denoising and inversion alternately by $K$ times, where each denoising step provides gradients from the target direction. Last, a pure successive denoising process is conducted for equal-step inversion and sampling.



From the manifold perspective, our ZigZag process can be regarded as performing iterative manifold constraint between the manifold of $p$ step (\ie,$\M_{p}$) and that of $p-1$ step (\ie,$\M_{p-1}$), where the target guidance is injected progressively to $\z_p$ while structure information of ${p}$ step is well preserved. Overall, our ZZEdit paradigm achieves better editing by achieving fewer fidelity errors. It can be painlessly applied to the existing methods which adopt the ``inversion-then-editing'' pipeline, and boost their performance. In Fig.~\ref{fig:teaser}, when our ZZEdit are equipped with P2P~\cite{p2p} and PnP~\cite{pnp}, more elegant editability and fidelity are achieved. Specifically, P2P supports DDIM inversion and Null-Text inversion (NTI)~\cite{null}, in which the latter delivers better results by optimizing unconditional textual embeddings. To sum up, our main contributions are:
\begin{itemize}
\item 
\vspace{-2mm}
We give a new empirical insight on using an intermediate-inverted latent $\z_p$ as editing pivot.
\vspace{-2mm}
\item
We propose a novel zero-shot image editing paradigm ZZEdit, where a ZigZag process performs iterative manifold constraint between the manifold $\M_{p}$ and $\M_{p-1}$, enhancing guidance at the pivot $\z_p$ mildly and decreasing accumulated fidelity errors.
\vspace{-2mm}
\item
Extensive qualitative and quantitative experiments demonstrate that our ZZEdit is versatile across different editing methods, including P2P~\cite{p2p} and PnP~\cite{pnp}, which achieves state-of-the-art editing performance.

\end{itemize}

\section{Related Works}

\noindent{\textbf{Text-driven Image Generation.}} Recent years, diffusion models~\citep{ddim,ddpm} has shown its capacity in text-to-image (T2I) generation.
DALLE-2~\citep{dalle2} proposes a two-stage model: a prior generating a CLIP~\citep{clip} image embedding given a text caption, and a decoder producing an image conditioned on the image embedding. 
Building on the strength of diffusion models in high-fidelity image generation, Imagen~\citep{imagen} discovers that large frozen language models trained only on text data are effective text encoders for text-to-image generation.
Further, to enable diffusion models training on limited computational resources while retaining quality, Stable Diffusion~\citep{ldm} trains 
models in the latent space of powerful pretrained autoencoders.

\noindent{\textbf{Text-driven Image Editing.}} 
Diffusion-based image editing modifies images with diffusion models using text instructions. SDEdit~\citep{sdedit} adds noise to the input (e.g., stroke painting), then subsequently
denoises through the prior from stochastic differential equation (SDE).
DiffusionCLIP~\citep{diffusionclip} proposes a text-guided image manipulation method using the pretrained diffusion models and CLIP loss.
To further improve the editing fidelity, some approaches require a mask region
~\citep{avrahami2023blended,avrahami2022blended,nichol2021glide}, where the background out of the mask can remain the same while it can be time-consuming for users to provide a mask. Then, for text-only intuitive image editing, DiffEdit~\citep{diffedit} and MasaCtrl~\citep{cao2023masactrl} automatically infer a mask according to the target prompt. P2P~\citep{p2p} and PnP~\citep{pnp} show that fine-grained control can be achieved by cross-attention layers and manipulating spatial features and their self-attention inside the model respectively. 
Besides, Imagic~\citep{imagic} and UniTune~\citep{valevski2022unitune} conduct fine-tuning on Imagen~\citep{imagen} to capture the image-specific appearance, which does not need edit masks either.
Further, InstructPix2Pix~\cite{brooks2023instructpix2pix} and MagicBrush~\cite{magicbrush} perform editing following instructions by constructing paired data. Pix2Pix-Zero~\citep{parmar2023zero} can perform image-to-image translation without manual prompting. Moreover, another line of techniques proposes to insert new concepts (e.g., a specified person, bag, cup) into a pretrained T2I model for personalize usage~\citep{dreambooth,tex_inv,dreamartist,kumari2023multi,smith2023continual,tewel2023key}.

\noindent{\textbf{Inversion in Editing Models.}}  
DDIM inversion~\citep{ddim} conducts DDIM sampling in the reverse direction, which is effective for unconditional generation. When the classifier-free guidance~\citep{cfg} is applied for conditional generation, the accumulated reconstruction error would magnify, thus bringing unsatisfied editing results. To address this, several methods~\citep{prompt,null} propose to perform optimization on inverted latents, where Null-text inversion (NTI)~\citep{null} optimizes the unconditional textual embedding while Prompt-Tuning inversion (PTI)~\citep{prompt} optimizes the conditional embedding.
There are also some techniques~\citep{ju2023direct,garibi2024renoise,edict} improve DDIM inversion without fine-tuning. 

This paper takes a close look at the latent trajectory of the existing ``inversion-then-editing'' pipeline, which we argue is usually suboptimal since it accumulates plenty of reconstruction errors. In contrast, we propose a new editing paradigm ZZEdit, which first locates a proper intermediate-inverted latent $\z_p$ with a better trade-off between \textit{editability} and \textit{fidelity}. Then, a ZigZag process is designed to mildly perform target guidance while holding structure information.

\section{Preliminary} 
\noindent{\textbf{Stable Diffusion (SD).}} 
SD~\citep{ldm} trains diffusion models for text-to-image in the latent space of an autoencoder $\mathcal{D}(\mathcal{E}(x))$. The encoder evaluates the latent feature $\z=\mathcal{E}(x)$ for an input image while the decoder $\mathcal{D}$ maps the latent representation to the RGB space. In the forward process, the latent input $\z_0$ is perturbed by Gaussian noise gradually, leading to $\z_T$. To sequentially denoising, a U-Net~\citep{unet} $\epsilon_{\theta}$ containing a series of residual, self-attention, and cross-attention blocks is trained to predict the noise by a L2 loss. Once trained, deterministic DDIM sampling~\citep{ddim} can accurately reconstruct a given real image using $\mathcal{C}$ as text embeddings: 
\begin{equation} 
  \begin{split}
 {\z}_{t\!-\!1}\!=\!\!\sqrt{\!\frac{\alpha_{t\!-\!1}}{\alpha_{t}}}{\z}_{t}\!+\!\!\sqrt{\alpha_{t\!-\!1}}\!\left(\!\!\sqrt{\!\frac{1}{\alpha_{t\!-\!1}}\!-\!1}\!-\!\!\sqrt{\!\frac{1}{\alpha_{t}}\!-\!1}\right)\!\!\epsilon_{\theta}(\!{\z}_{t},\!t,\!\C\!),
  \end{split}
\label{ddim_sam}
\end{equation} 
\noindent{\textbf{DDIM Inversion.}} 
{DDIM inversion}~\cite{ddim} projects an image into a known latent space for editing, which performs DDIM sampling process in a reverse way:
\begin{equation}
{\z}_{t}\!\!=\!\!\sqrt{\!\frac{\alpha_{t}}{\alpha_{t\!-\!1}}}{\z}_{t\!-\!1}+\sqrt{\alpha_{t}} \left(\!\!\!\sqrt{\!\frac{1}{\alpha_{t}}\!-\!1}\!-\!\!\sqrt{\!\frac{1}{\alpha_{t\!-\!1}}\!-\!1}\right)\!\!\epsilon_{\theta}(\!{\z}_{t\!-\!1},\!t\!-\!1,\!\C\!).
\label{ddim_inv}
\end{equation}
The technique is based on the assumption that the ODE process can be reversed in the limit of small steps. 

\noindent{\textbf{Classifier-free Guidance (CFG).}}
To enhance the guidance of the text condition in text-driven generation, classifier-free guidance~\citep{cfg} is proposed, where conditional and unconditional prediction are combined at each step. The calculation is defined as:
\begin{equation}
\Tilde{\epsilon}_{\theta}(\z_t, t, \C, \varnothing) = \omega \cdot  \epsilon_{\theta}(\z_t, t, \C) + (1-\omega) \cdot \epsilon_{\theta}(\z_t, t, \varnothing),
\label{cfg}
\end{equation}
where $\varnothing$ is the embeddings of a null text, and $\omega$ is the guidance scale parameter. Note that a slight error is introduced in each step of DDIM inversion, and popular usage of large guidance scale $\omega>1$ would magnify such accumulated errors~\cite{null,prompt}.



\begin{figure*}[t]
    \centering
    \includegraphics[width=1\textwidth]{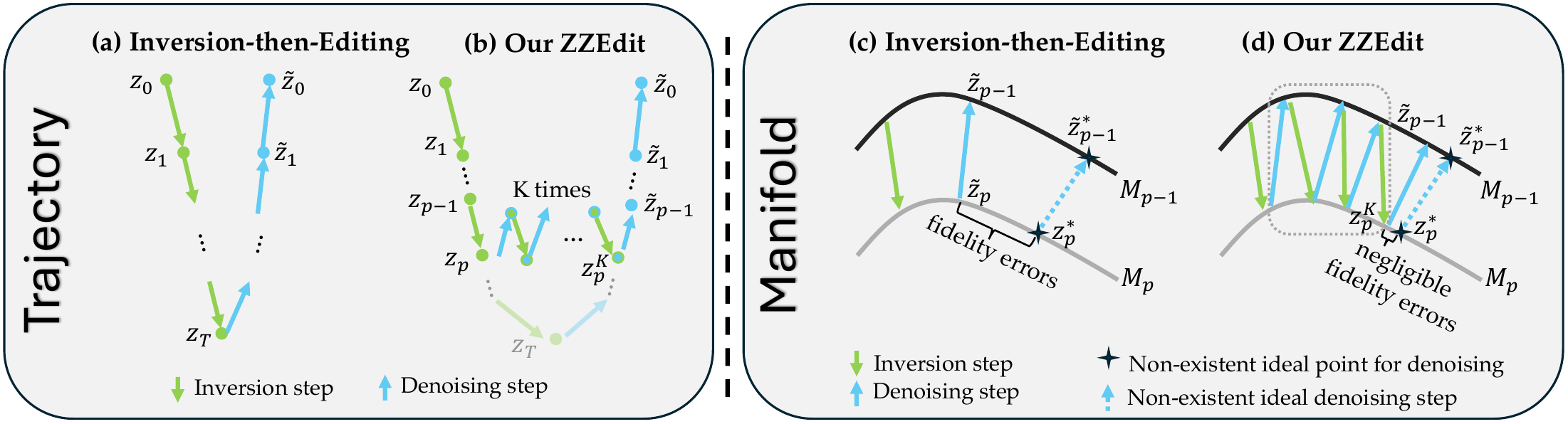}
    \vspace{-2em}
    \caption{Left: The trajectory of the ``inversion-then-editing'' pipeline and our ZZEdit. (a) The former invertes $\z_0$ to $\z_T$ using $\mathcal{P}_{src}$, and then carry out denoising under $\mathcal{P}_{tgt}$. (b) The latter first locates a qualified intermediate-inverted latent marked as $\z_p$ as a better editing pivot, which is sufficient-for-editing while structure-preserving. Then, a ZigZag process is proposed to mildly perform target guidance by alternately executing one-step denoising and inversion by $K$ times. Afterwards, a pure denoising process is leveraged for the equal step of inversion and denoising. Right: Manifold illustration of ``inversion-then-editing'' pipeline and our ZZEdit at the step $p$ and $p-1$. (c) The former shows noticeable fidelity lost between the denoised latent $\Tilde{\z}_p$ and the ideal one ${\z}_p^*$ when reconstructing semantics from a noisy manifold $\M_T$. (d) The latter conducts iterative manifold constraint on ${\z}_p$, to which target guidance is progressively injected without ruining the structure information of ${\z}_p$. The corresponding ${\z}_p^K$ is closer to the optimal point $\Tilde{\z}_p^*$ for the next pure denoising process.}
    \label{fig:pipeline}
       \vspace{-1em}
\end{figure*}
\section{Methods}
\label{sec:method} 
\subsection{Pilot Analysis}
\label{sec:problem}
Given a source image $I$ and a target prompt $\mathcal{P}_{tgt}$, text-driven image editing tries to achieve two needs: \textbf{\textit{editability}} and \textbf{\textit{fidelity}}. The former aims to change visual content to be consistent with the textual description of $\mathcal{P}_{tgt}$, while the latter requires the rest to remain unchanged. As shown in Fig.~\ref{fig:pipeline} (a), recent text-only image editing methods always obey the ``inversion-then-editing'' pipeline. It inverts the input image embedding $\z_0$ (on the source manifold $\M^{src}_0$) for $T$ steps to obtain an approximately standard Gaussian noise $\z_T$ (on the noisy manifold $\M_{T}$), from which a sampling process is conducted under the target prompt $\mathcal{P}_{tgt}$ using CFG. However, \textit{we argue that it is not a good choice to directly invert the input image to a near-Gaussian noise.} Next, we leverage DDIM inversion to verify this from the perspective of the trade-off between \textit{editability} and \textit{fidelity}.

\begin{figure}[t]
    \centering
    \includegraphics[width=0.48\textwidth]{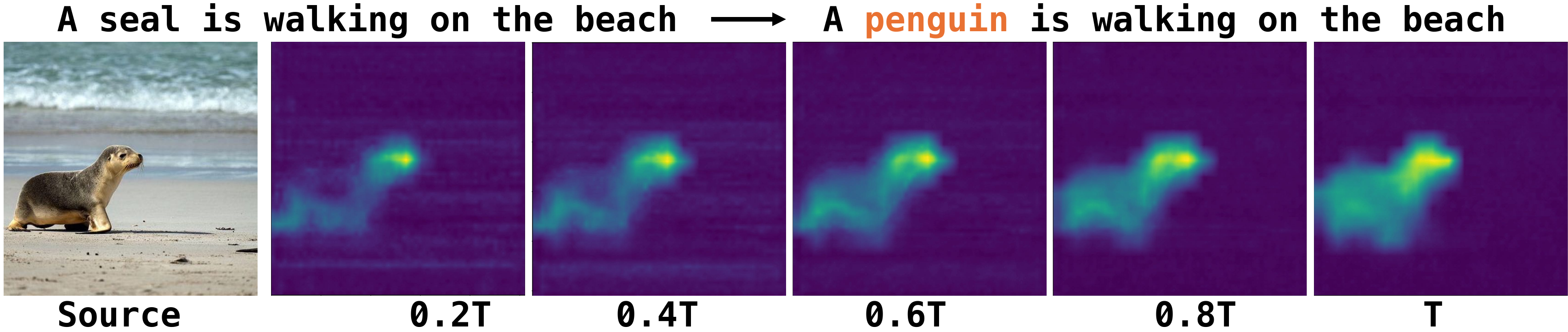}
    \vspace{-2.2em}
    \caption{The cross-attention maps between different inverted latents $\z_t$ and the target prompt $\mathcal{P}_{tgt}$.
    }
    \label{fig:intro}
    \vspace{-1.2em}
\end{figure}

\noindent{\textbf{Editability.}}
We use cross-attention maps to reflect the editability of different inverted latent $\z_{t}$ towards the target prompt $\mathcal{P}_{tgt}$. For brevity, we divide the T-step process into five parts, where $t \in [0.2T, 0.4T, 0.6T, 0.8T, T]$. In Fig.\ref{fig:intro}, we show an example of ``a \sout{\textit{seal}} \textit{penguin} walking on the beach'', where intermediate-inverted latents (\eg, $t=0.6T$ and $t=0.8T$) can provide considerable response level to the target prompt $\mathcal{P}_{tgt}$ as the fully-inverted $\z_T$. Generally speaking, we think that target prompt $\mathcal{P}_{tgt}$ usually has some shared semantics (\eg, \textit{beach}) with the source image, and these semantics do not need to be perturbed completely for reconstruction latter. Intermediate-inverted latents always have good potential to deliver sufficient editability for those to-be-edited contents. More visualizations are in Appendix.

\noindent{\textbf{Fidelity.}} 
DDIM inversion introduces a slight error at each step, and such accumulated errors would be magnified under a large CFG scale $\omega$~\cite{null}. Thus, using $\z_T$ for subsequent denoising would bring more reconstruction errors than intermediate-inverted ones, hindering the fidelity and sometimes leading to a totally different image. 
Summing up, intermediate-inverted latents can give a better trade-off between \textit{editability} and \textit{fidelity} than $\z_T$.


\subsection{Overview of The Proposed ZZEdit}
\label{sec:zzedit}
Given the above pilot analysis, this paper proposes a new editing paradigm named ZZEdit, which \textit{mildly strengthens the target guidance on a sufficient-for-editing while structure-preserving latent.} Specifically, as seen in Fig.~\ref{fig:pipeline} (b), our ZZEdit  consists of three parts:\\ 
{(i) We locate a proper intermediate-inverted latent marked as $\z_p$ as a better editing pivot, which is in Sec.~\ref{sec:pivot}.} \\
{(ii) A ZigZag process is proposed, which alternately executes one-step denoising and inversion by $K$ times to mildly enhance target guidance. Concretely, it fulfills iterative manifold constraint between the manifold of $p$ step ($\M_{p}$) and that of $p-1$ step ($\M_{p-1}$), which is elaborated in Sec.~\ref{sec:zigzag}. }\\
{(iii) The remaining comprises a diffusion process guided by the target prompt $\mathcal{P}_{tgt}$ to achieve equal-step inversion and sampling. Note that when equipping the existing editing method with our ZZEdit, the denoising process needs to retain the characteristics of the method, such as P2P~\citep{p2p} injecting cross-attention maps and PnP~\citep{pnp} injecting self-attention maps. }
We summarize applying our ZZEdit to the existing text-driven image editing methods in Alg.~\ref{alg:pivot}.

\subsection{Locating a Better Pivot on Inversion Trajectory}
\label{sec:pivot}
We seek a qualified intermediate-inverted latent marked as $\z_p$ as editing pivot, which considers both \textit{editability} and \textit{fidelity}. Editability is achieved by locating a sufficient-for-editing point which has a larger response towards the target prompt $\mathcal{P}_{tgt}$ than the source $\mathcal{P}_{src}$. Fidelity is naturally guaranteed by fewer inverted steps than $T$.
Specifically, we apply the response of the pretrained U-Net $\epsilon_{\theta}$ to locate such a pivot. Starting from $t=1$, given an inverted latent $\z_t \in \{\z_1,...,\z_T\}$, we use Eqn.~\ref{ddim_sam} for one-step DDIM sampling, obtaining the denoised latent $\hat{\z}_{t-1}$, $\bar{\z}_{t-1}$, and $\Tilde{\z}_{t-1}$ under the source prompt $\mathcal{P}_{src}$, null text $\varnothing$, and target prompt $\mathcal{P}_{tgt}$:
\begin{equation}
\label{eq:responce}
\begin{split}
\vspace{-0.05em}
\hat{\z}_{t-1}\!\!\gets\!\!\epsilon_{\theta}(\z_t,\! t, \!\C_{src}),  
\Tilde{\z}_{t-1}\!\!\gets\!\!\epsilon_{\theta}(\z_t,\! t,\! \C_{tgt}),
\bar{\z}_{t-1}\!\!\gets\!\!\epsilon_{\theta}(\z_t,\! t, \!\varnothing).
\nonumber
\vspace{-0.05em}
\end{split}
\end{equation}
Then, we measure the response level towards the target prompt $\mathcal{P}_{tgt}$ as $\|\Tilde{\z}_{t-1} - \bar{\z}_{t-1}\|$ and that towards the source prompt $\mathcal{P}_{tgt}$ as $\|\hat{\z}_{t-1} - \bar{\z}_{t-1}\|$. 
Generally speaking, the latent $\z_t$ with low-degree inversion would be more responsive to source prompt $\mathcal{P}_{src}$ due to limited corruption. As the inversion deepens, we can easily find those points whose response to $\mathcal{P}_{tgt}$ is greater than that to $\mathcal{P}_{src}$: 
\begin{equation}
\label{eq:greater_responce}
\begin{split}
\vspace{-0.05em}
\|\Tilde{\z}_{t-1} - \bar{\z}_{t-1}\| > \|\hat{\z}_{t-1} - \bar{\z}_{t-1}\|.
\end{split}
\vspace{-0.05em}
\end{equation}
Here, the denoised latent $\bar{\z}_{t-1}$ with $\varnothing$ is used as an anchor. The more the denoised latent deviates from that of ${\varnothing}$, the greater the response.
The intuition is that when an intermediate-inverted ${\z}_t$ can deliver a larger response towards $\mathcal{P}_{tgt}$ from U-Net $\epsilon_{\theta}$, we believe it is a sufficient-for-editing point.
For simplicity, we only locate the \textit{first} point during inversion which has a larger target response as our editing pivot. We mark the satisfied step $t$ as $p \in[1,...,T]$. 

\subsection{Iterative Manifold Constraint: ZigZag Process}
\label{sec:zigzag}
To mildly deepen editing without ruining the fidelity of previously located pivot $\z_p$, we propose a ZigZag process, which alternately executes one-step sampling and inversion.

\begin{algorithm}[t]
   \caption{ZZEdit for Zero-shot Image Editing}
\begin{algorithmic}[1]
 \newcommand{\algrule}[1][.9pt]{\par\vskip.1\baselineskip\hrule height #1\par\vskip.1\baselineskip}
   \STATE {\bfseries Input:} The inverted latents $\{\z_1,...,\z_T\}$, source prompt $\mathcal{P}_{src}$, and target prompt $\mathcal{P}_{tgt}$
   \STATE {\bfseries Output:} {An edited image or latent embedding ${\tilde{\z}}_0$}
\algrule
Part I: Seeking a Better Pivot $p$ on Inversion Trajectory
\algrule
   \FOR{$t=1$ {\bfseries to} $T$}  
\STATE $\hat{\z}_{t-1} \gets \epsilon_{\theta}(\z_t,t,\C_{src})$; \quad $\triangleright$ {Eqn.~\ref{ddim_sam} with $\C_{src}$}
\STATE $\Tilde{\z}_{t-1} \gets \epsilon_{\theta}(\z_t,t, \C_{tgt})$; \quad $\triangleright$ {Eqn.~\ref{ddim_sam} with $\C_{tgt}$}
\STATE $\bar{\z}_{t-1} \gets \epsilon_{\theta}(\z_t, t, \varnothing)$; \quad $\triangleright$ {Eqn.~\ref{ddim_sam} with $\varnothing$}
   \IF {$\|\Tilde{\z}_{t-1} - \bar{\z}_{t-1}\| > \|\hat{\z}_{t-1} - \bar{\z}_{t-1}\|$}
        \STATE \textbf{break}
        \ENDIF
         \ENDFOR
\RETURN $t$;
\algrule
\noindent{Part II: Iterative Manifold Constraint: ZigZag Process}
\algrule
\FOR{$k=1$ {\bfseries to} $K$} 
\STATE $\Tilde{\z}^{k}_{p-1}\gets \epsilon_{\theta}(\z^{k-1}_{p},p,\C_{tgt})$; \quad $\triangleright$ {Eqn.~\ref{eq:den} at $k$-th union}
\STATE $\z^k_{p} \gets \epsilon_{\theta}(\z^k_{p-1},p-1,\C_{src})$; \quad $\triangleright$ {Eqn.~\ref{eq:inv} at $k$-th union}
   \ENDFOR
\algrule
Part III: Continuous Denoising Process
\algrule
\FOR{$t=p$ {\bfseries to} $1$}
{\STATE $\Tilde{\z}_{t\!-\!1}\!\!\gets\!\!\epsilon_{\theta}(\!\z_t,\!t,\!\C_{tgt}\!)$; \quad
 $\triangleright$ {Eqn.~\ref{ddim_sam} with P2P or PnP}}
 \ENDFOR
\end{algorithmic}
\label{alg:pivot}
\end{algorithm}

\noindent{\textbf{Mild Guidance.}} As illustrated in Fig.\ref{fig:pipeline} (b), our ZigZag process is started after a $p$-step inversion.
Formally, a full ZigZag process includes $K$ denoising steps and $K$ inversion steps, which are conducted alternately. We treat a  
denoising step and inversion step as a union, making ZigZag process consist of $K$ unions. The inversion step of $k$-th union is:
\begin{equation} 
\begin{split}
\label{eq:inv}
\z^k_{p}\!=\!\!\sqrt{\!\frac{\alpha_{p}}{\alpha_{p\!-\!1}}} \z^k_{p-1} &\!+\!\sqrt{\alpha_{p}}\!\left(\!\!\sqrt{\!\frac{1}{\alpha_{p}}\!-\!1}\!-\!\!\sqrt{\!\frac{1}{\alpha_{p\!-\!1}}\!-\!1}\right)\!\!\epsilon^{p\!-\!1}_\theta, \\
\epsilon^{p\!-\!1}_\theta&=\epsilon_{\theta}(\z^k_{p-1},p-1,\C_{src}),
\end{split}
\end{equation}
where $k\in\{1,2,...,K\}$.
Then, the denoising step of $(k+1)$-th union in ZigZag process is:
\begin{equation}
\label{eq:den}
\begin{split}
    \Tilde{\z}^{k+1}_{p-1}\!=\!\!\!\sqrt{\!\frac{\alpha_{p-1}}{\alpha_{p}}} 
    \z^k_{p}&\!+\!\!\sqrt{\alpha_{p\!-\!1}}\!\left(\!\!\sqrt{\!\frac{1}{\alpha_{p\!-\!1}}\!-\!1}\!-\!\!\sqrt{\!\frac{1}{\alpha_{p}}\!-\!1}\right)\!\!\epsilon_{\theta}^{p}, \\
    \epsilon_{\theta}^{p}&=\epsilon_{\theta}(\z^k_{p},p,\C_{tgt})
\end{split}
\end{equation}
Then, we can re-write the denoised latent of $(k+1)$-th union in ZigZag process by substituting Eqn.~\ref{eq:inv} into Eqn.~\ref{eq:den}:
\begin{equation}
\label{eq:itr}
\begin{split}
\Tilde{\z}^{k+1}_{p-1}&\!=\!\z^k_{p\!-\!1}\!+\!\sqrt{\alpha_{p\!-\!1}} \left(\!\!\sqrt{\frac{1}{\alpha_{p\!-\!1}}\!-\!1}\!-\sqrt{\frac{1}{\alpha_{p}}\!-\!1}\right)\!\Delta\epsilon_{\theta},
\\
\Delta\epsilon_{\theta}&=\epsilon_{\theta}^{p}\!-\!\epsilon^{p\!-\!1}_\theta
\end{split}
\end{equation}
where $\sqrt{\alpha_{p\!-\!1}} \left(\!\sqrt{\frac{1}{\alpha_{p-1}}-1}\!-\sqrt{\frac{1}{\alpha_{p}}-1}\right)>0$ according to noise schedule of diffusion models~\citep{ddim}. Thus, the denoised latent in $(k+1)$-th union (\ie, $\Tilde{\z}^{k+1}_{p-1}$) would move towards target compared with that in $k$-th union (\ie, $\Tilde{\z}^{k}_{p-1}$). Overall, our ZigZag process progressively injects target guidance into located pivot $\z_p$, while the structure information of $p$ step is well preserved.


\begin{figure*}[t]
    \centering
    \includegraphics[width=1\textwidth]{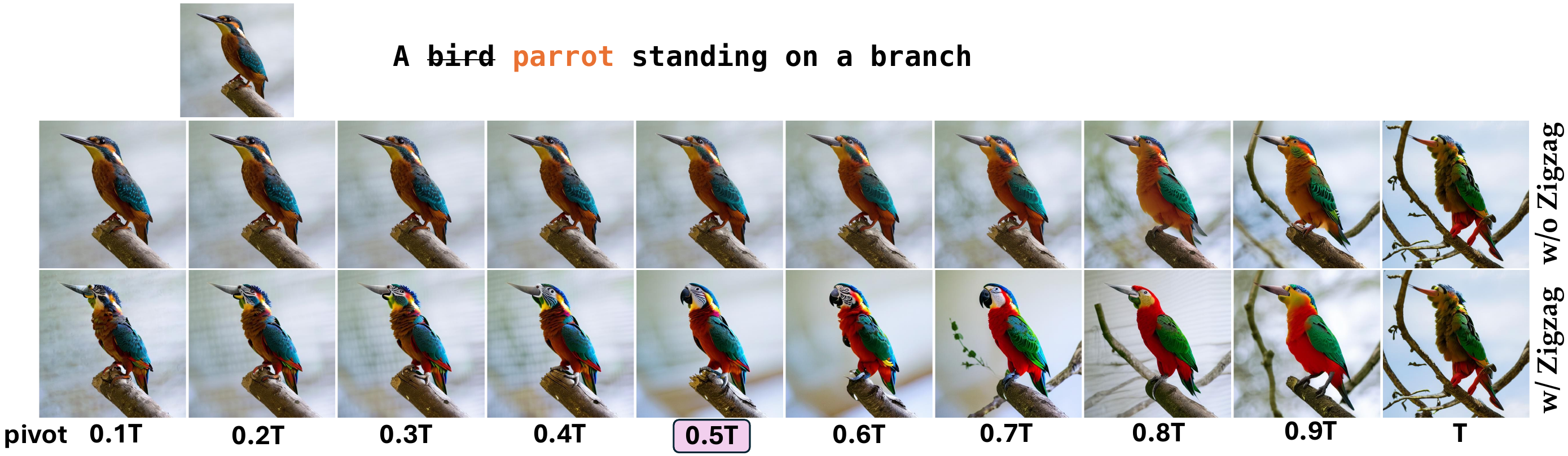}
    \vspace{-2em}
    \caption{Ablation study of ZZEdit on P2P~\citep{p2p} \textit{w/} DDIM inversion. The first row displays the results of using different inverted $\z_t$ as editing pivot without ZigZag process. The second row shows the performance of using the ZigZag process additionally. Our method first locates a suitable pivot $\z_{p}$ (marked with purple) and then mildly performs target guidance, yielding the most elegant results.}
    \label{fig:ab_diff}
\end{figure*}


\begin{table*}[t]
    \centering
    \small
\begin{tabular}{c|l|c|c|c|c|c|c|c}
  \toprule[1pt]
\multicolumn{2}{c|}{\multirow{2}{*}{\textbf {Method}}} &\textbf{Structure} & \multicolumn{4}{c|}{\textbf{Background Preservation}} &\multicolumn{2}{c}{\textbf{CLIP Similariy}}\\
\cline{3-9}

\multicolumn{2}{c|}{} & \textbf{L2} $\downarrow$  &\textbf{PSNR}$\uparrow$ &\textbf{LPIPS} $\downarrow$ &\textbf{MSE} $\downarrow$ &\textbf{SSIM} $\uparrow$  &\textbf{Whole}$\uparrow$ &\textbf{Edited}$\uparrow$  \\
\hline

\multicolumn{2}{c|}{\textbf{P2P+DDIM Baseline}}  &69.41  &17.88 &208.37 &219.11 &71.30 &25.01 &{22.44}  \\ 
\hline
\multirow{4}{*}{\bm{$w/$}  \textbf{Pivot}}
 &\bm{$w/o$} \textbf{ZigZag} (\bm{$a=0$})  &\textbf{22.60}  &\textbf{23.71} &\textbf{107.01} &\textbf{68.27} &\textbf{79.60} &{24.43} &{21.52} \\ 
 &\bm{$w/$} \textbf{ZigZag} (\bm{$a=0.2$})  &27.50  &22.97 &116.02 &82.79 &78.71 &24.70 &22.04 \\ 
 &\bm{$w/$} \textbf{ZigZag} (\bm{$a=0.6$})    &28.26  &22.48 &122.36 &87.26 &77.94 &25.07 &22.14 \\
 &\bm{$w/$} \textbf{ZigZag} (\bm{$a=1$})     &31.99  &21.92 &131.57 &96.95 &76.98 &\textbf{25.29} &\textbf{22.47}\\
 \hline
\multicolumn{2}{c|}{\textbf{Random Pivot} \bm{$w/$} \textbf{ZigZag} (\bm{$a=1$})}  &{25.84}  &{24.07} &{105.36} &{81.43} &{79.56} &24.76 &21.84   \\ 
\bottomrule[1pt] 
\end{tabular}
\vspace{-0.8em}
\caption{Quantitative ablation study on our ZigZag process with P2P~\citep{p2p} \textit{w/} DDIM inversion. We mark the best results of 
ZZEdit using located pivot $\z_p$ in bold. We also provide the performance of random editing pivot with a standard ZigZag process. }
\label{tab:ablation}
\vspace{-0.6em}
\end{table*}



\noindent{\textbf{Manifold Perspective for ZigZag Process.}}
To further demonstrate the role of our ZigZag process, we dive into the manifold of $p$ and $p-1$ steps. In Fig.~\ref{fig:pipeline} (c), since DDIM inversion introduces a slight error in each step, 
the typical ``inversion-then-editing'' pipeline has noticeable fidelity errors between the denoised point $\Tilde{\z}_p$ and the ideal one $\z_p^*$ when reconstructing semantics from a noisy manifold $\M_T$. Here, a large guidance scale $\omega>1$ of CFG magnifies such accumulated errors~\cite{null,prompt}.

As seen in Fig.~\ref{fig:pipeline} (d), our ZigZag process performs iterative manifold constraint between the manifold of $p$ step ($\M_{p}$) and that of $p-1$ step ($\M_{p-1}$), where ${\z}_p$ itself shows a better trade-off between \textit{editability} and \textit{fidelity}, to which target guidance is injected progressively. Here, although each inversion step in our ZigZag process also introduces slight errors, the structure information still can be maintained since instant denoising follows each inversion step, avoiding larger errors being accumulated by the noise schedule of diffusion models~\citep{ddim}. Thus, the output of our ZigZag process (\ie, ${\z}_p^K$) is closer to the non-existent optimal point ${\z}_p^*$ for the next pure denoising process. Generally, our ZZEdit achieves better editing than the ``inversion-then-editing'' pipeline by decreasing fidelity errors.



\noindent{\textbf{ZigZag Steps.}} For a fair comparison, we use the same steps of inversion and sampling with the typical ``inversion-then-editing'' pipeline to determine ZigZag steps, where $p+K=T$. Then, when $p=T$, ZZEdit would degenerate to the typical ``inversion-then-editing'' pipeline. Besides, we additionally introduce a hyper-parameter $a\in[0,1]$ as:
\begin{equation}
K = a \cdot (T-p),
\label{zz_steps}
\end{equation}
where $a$ can control ZigZag steps flexibly. When $a=0$, a continuous $p$-step sampling is performed from the located editing pivot ${\z}_p$ without ZigZag process. When $a=1$, our ZZEdit realizes $T$ inversion and sampling steps separately, consuming the same UNet operations as the ``inversion-then-editing'' pipeline. More discussion of additional UNet operations during locating pivot ${\z}_p$ is in Appendix.

\section{Experiment}
\label{sec:exp}
\subsection{Experimental setup}
\noindent{\textbf{{Implementation Details.}} All experiments are conducted on a single Tesla A100 GPU using PyTorch~\cite{paszke2019pytorch}. Following~\cite{pnp}, we use 50 steps as DDIM schedule and the classifier-free guidance of 7.5 for editing. Besides, we use the official code of SD 1.5.
For a fair comparison, we adopt the same cross-attention injection parameters and self-attention injection parameters as P2P~\cite{p2p} and PnP~\cite{pnp}. In practice, to save time and computation, when looking for the editing pivot, we only search from $[0.4T, 0.5T,..., T]$, rather than $[0, 1,..., T]$. The reasons are: (1) low-degree inversion generally struggles for sufficient editability and (2) there is no need to look up each step.


\noindent{\textbf{{Evaluation Metrics.}}
We use the PIE-Bench dataset~\citep{ju2023direct} to evaluate our method. The editing results are evaluated on three aspects: structure distance~\citep{tumanyan2022splicing},  background preservation covering PSNR~\citep{psnr}, SSIM~\citep{ssim}, MSE, and LPIPS~\citep{lpips}, and editing consistency of the whole image and regions in the
editing mask, denoted as CLIP similarity~\citep{wu2021godiva}.



\subsection{Ablation Studies}
We ablate several key designs of our ZZEdit paradigm, which aims to answer the following questions. 
\textbf{Q1:} {What is the difference between using different points on the inversion trajectory as editing pivot?}
\textbf{Q2:} {Could our ZZEdit locate a suitable pivot $\z_p$, which maintains both fidelity and editability?}
\textbf{Q3:} {Could the proposed ZigZag process enhance the target guidance at the suitable pivot $\z_p$?}

\begin{table*}[t]
  \centering
  \small
\begin{tabular}{l|c|c|c|c|c|c|c|c}
  \toprule[1pt]
  \multicolumn{2}{c|}{\textbf{Method}} & \textbf{Structure}   &\multicolumn{4}{c|}{\textbf{Background Preservation}} &\multicolumn{2}{c}{\textbf{CLIP Similariy}}
  \\\hline
  \textbf{Editing}      & \textbf{Inv Setting}  &\textbf{L2} $\downarrow$  &\textbf{PSNR}$\uparrow$ &\textbf{LPIPS} $\downarrow$ &\textbf{MSE} $\downarrow$ &\textbf{SSIM} $\uparrow$  &\textbf{Whole}$\uparrow$ &\textbf{Edited}$\uparrow$
  \\\hline
  \multirow{5}{*}{\textbf{P2P}}       & \textbf{DDIM}    &69.41  &17.88 &208.37 &219.11 &71.30 &25.01 &22.44\\ 
                             & \textbf{NTI}     &13.72 &27.05 &60.74 &35.89&84.27 &24.75 &21.86\\ 
                             &\textbf{PTI}   &16.17  &26.21 &69.01 &39.73 &83.40 &24.61 &21.87\\ 
                             &\textbf{Pnp$\_$inv}     & 
                             {11.65} &{27.22}&{54.55}&{32.86}&{84.76} &25.02&22.10\\
                             \cline{2-9}
                    & \textbf{ZZEdit (\bm{$w/$} DDIM)}    &31.99  &21.92 &131.57 &96.95 &76.98 &\textbf{25.29} &\textbf{22.47}\\
                    & \textbf{ZZEdit (\bm{$w/$} NTI)}   &\textbf{11.47} &\textbf{27.42} &\textbf{53.92} &\textbf{31.23} &\textbf{84.98} &24.95 &22.01 \\

                             \hline
                              \hline
  \multirow{3}{*}{\textbf{PnP}}         
  & \textbf{DDIM}     &28.22  &22.28&113.46&83.64& 79.05&25.41& 22.55\\ 
&\textbf{Pnp$\_$inv}     &24.29  &22.46 &106.06&80.45& 79.68&25.41&22.62 \\  
\cline{2-9}
& \textbf{ZZEdit (\bm{$w/$} DDIM)}     &\textbf{23.49}  &\textbf{24.55} &\textbf{86.61} &\textbf{55.04} &\textbf{82.18} &\textbf{25.43} &\textbf{22.91}\\
\bottomrule[1pt] 
\end{tabular}
\vspace{-0.8em}
  \caption{
  Comparison between ZZEdit and ``inversion-then-editing'' pipeline on P2P~\citep{p2p} and PnP~\citep{pnp} on different inversion settings: DDIM~\cite{ddim}, NTI~\citep{null}, PTI~\citep{prompt}, and Pnp~\citep{ju2023direct}.
  }
  \label{tab:quantitative}
\vspace{-0.4em}
\end{table*}





\noindent{\textbf{\noindent{\textbf{Different Editing Pivots in ZZEdit.}}}} In Fig.~\ref{fig:ab_diff}, we answer the first question by applying ZZEdit on P2P~\citep{p2p} \textit{w/} DDIM inversion and report the performance of selecting the value of $p$ from $[0.1T, 0.2T, ...,0.9T, T]$ as different editing pivots.
The first row uses $p$-step inversion and $p$-step sampling without the ZigZag process. The second row displays the results of ZigZag process equipped for different inverted latents, where each result in the second row satisfies $p+K=T$ to make UNet operations the same as the ``inversion-then-editing'' baseline.

From the first row, we can observe that different inverted latents have different response levels to the target prompt. When we choose $[0.1T, 0.2T,0.3T,0.4T]$ as editing pivot, structure fidelity is maintained well, but editability is poor. It demonstrates that a low-degree inversion struggles to bring sufficient editability. Besides, we notice that when using high-degree inversion (e.g.,$[0.8T, 0.9T,T]$), the corresponding results deliver satisfactory editability but an unpleasing background since plentiful accumulated fidelity errors are introduced during reconstruction. For the second row, we equip ZigZag process at different inverted latents. Note that using ZigZag process for those low-degree inverted latents shows limited editing consistency since it only performs mild guidance on these latents, where structure information is preserved. Fortunately, our method first finds a sufficient-for-editing while structure-preserving point $z_p$ (marked in purple), and then performs mild guidance, which yields the most elegant performance. We also use GPT-4V(ision) system~\citep{gpt4-v} to evaluate Fig.~\ref{fig:ab_diff} in Appendix.

\noindent{\textbf{The Effectiveness of Our Located Pivot.}} 
We answer the second question by comparing the results of selecting editing pivot randomly from $[0.1T, 0.2T,...,0.9T,T]$ when a standard ZigZag process ($a=1$) is equipped.
As shown in Tab.~\ref{tab:ablation}, compared with P2P baseline, although ``random pivot \textit{w/} ZigZag'' can achieve more excellent background and structure preservation, its editing consistency is poor. The reason is that when the randomly selected pivot is low-degree inverted latent, our ZigZag process brings limited target guidance. In contrast, our standard ZZEdit achieves much higher CLIP scores, which proves its effectiveness of locating a sufficient-for-editing while structure-preserving intermediate-inverted latent $\z_p$ as editing pivot. Besides, we also give the distribution of our located editing pivots on the  PIE-Bench dataset~\citep{ju2023direct} in Appendix.


\begin{figure}[t]
    \centering
    \includegraphics[width=0.49\textwidth]{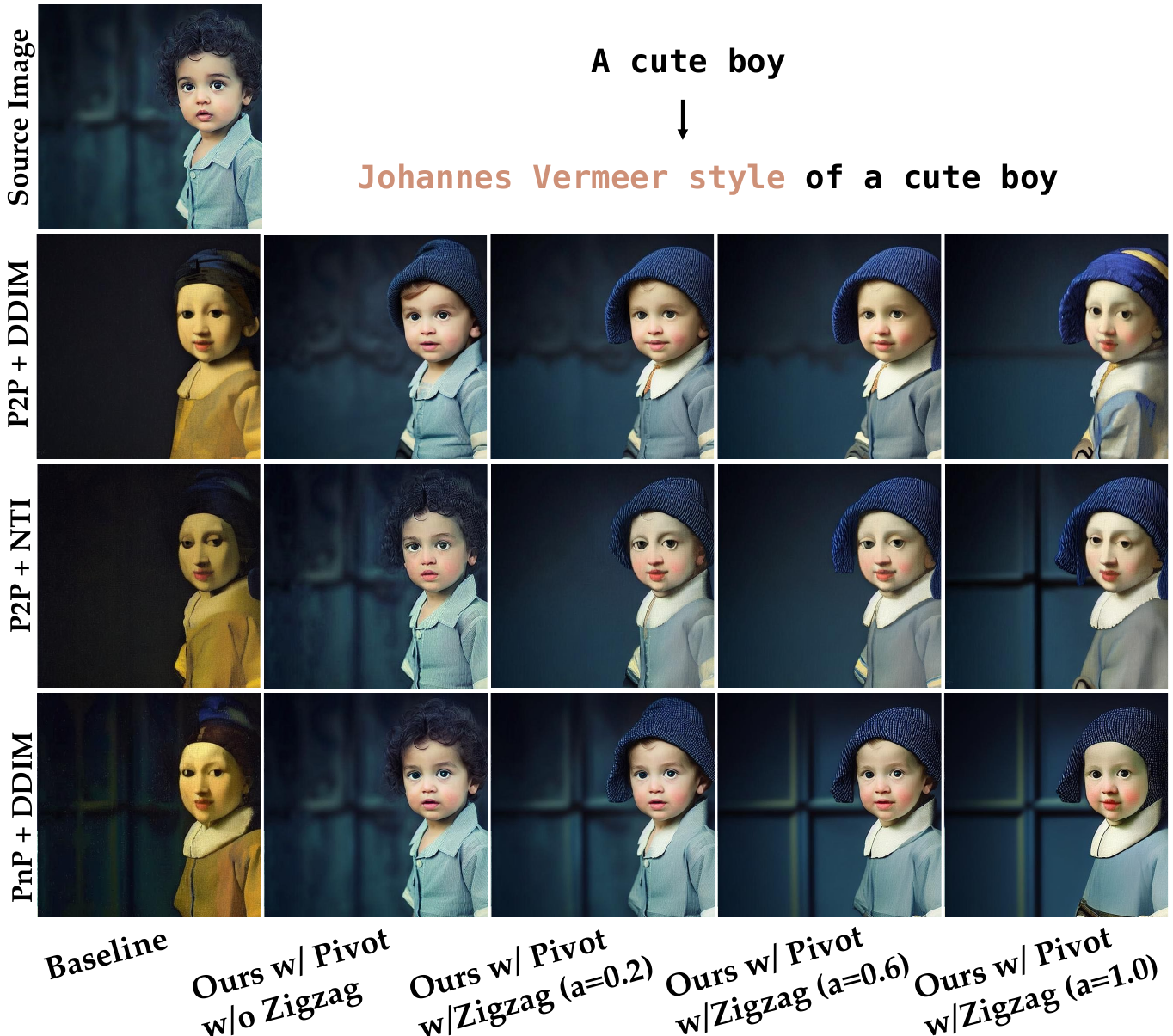}
    \vspace{-1.8em}
    \caption{Qualitative ablation on our ZigZag process with P2P~\citep{p2p} and PnP~\citep{pnp}, which mildly enhances the guidance at a suitable pivot $\z_{p}$.
    }
    \label{fig:ablation}
    \vspace{-0.5em}
\end{figure}

\noindent{\textbf{The Effectiveness of The ZigZag Process.}}
We answer the third question by using different ZigZag steps on a suitable editing pivot $\z_p$, which makes $a$ in Eq.~\ref{zz_steps} take the value from $\{0, 0.2, 0.6, 1\}$. 
Fig.~\ref{fig:ablation} shows a qualitative comparison on different baselines. Our ZigZag process can progressively inject target guidance through the increasing number of ZigZag steps while still holding a satisfying background. We also provide a quantitative experiment in Tab.~\ref{tab:ablation}. When no ZigZag steps are employed ($a=0$), the best background and structure can be obtained. However, it cannot achieve pleasing editing consistency. Besides, the gradual increase of ZigZag steps ($a=0.2, 0.6$, and $1$) can effectively improve editing consistency. 
Here, the structure and background are slightly weakened but still at a desirable level. The quantitative ablations on the ZigZag process with P2P \textit{w/} NTI and PnP \textit{w/} DDIM inversion are in Appendix.


\begin{figure*}[t]
    \centering
    \includegraphics[width=1\textwidth]{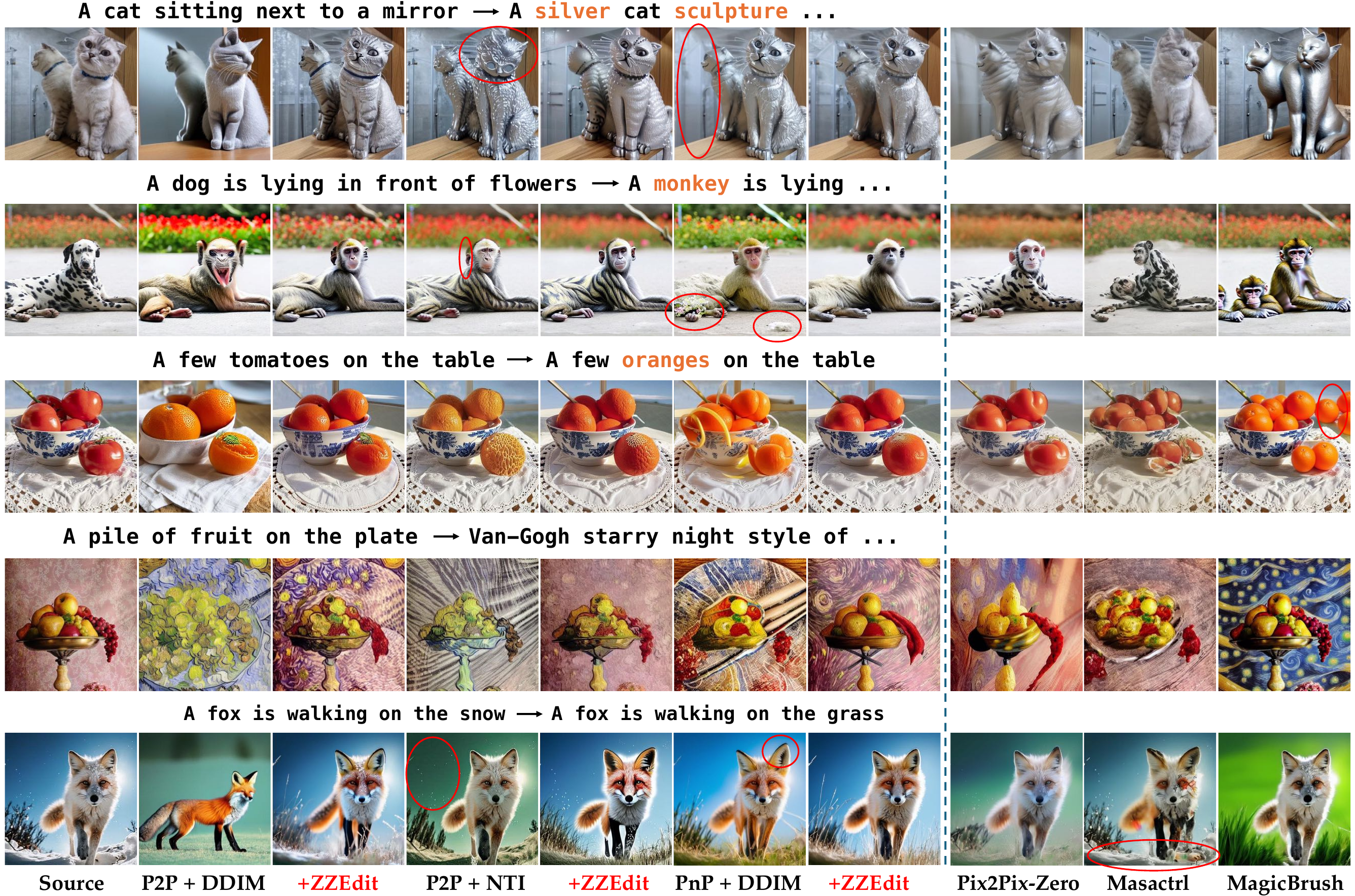}
    \vspace{-2em}
\caption{Visualization results of different editing techniques. From left to right: source image, P2P~\citep{p2p} \textit{w/} DDIM inversion, our ZZEdit applied on P2P \textit{w/} DDIM inversion, P2P \textit{w/} Null-text inversion, our ZZEdit applied on P2P \textit{w/} Null-text inversion, PnP~\citep{pnp} \textit{w/} DDIM inversion, our ZZEdit applied on PnP \textit{w/} DDIM inversion, Pix2Pix-Zero~\citep{parmar2023zero}, Masactrl~\citep{cao2023masactrl}, MagicBrush~\citep{magicbrush}.}
    \label{fig:qualitative}
    \vspace{-1em}
\end{figure*}
\subsection{Quantitative Results}
To prove the superiority of our ZZEdit, we compare it with P2P~\citep{p2p} and PnP~\citep{pnp} under different inversion settings.
As seen in Tab.~\ref{tab:quantitative}, when applying ZZEdit to P2P or PnP, all results of background, structure, and editing consistency are boosted steadily. Besides,
PnP \textit{w/} ZZEdit outperforms the PnP \textit{w/} Pnp inversion clearly. Further, P2P + NTI \textit{w/} ZZEdit yields a comparable performance with P2P \textit{w/} Pnp inversion~\citep{ju2023direct}.



\subsection{Qualitative Results}
In Fig.~\ref{fig:qualitative}, we show a qualitative comparison with the current editing methods, including P2P~\cite{p2p} \textit{w/} DDIM inversion or NTI, PnP~\cite{pnp} \textit{w/} DDIM inversion, Pix2Pix-Zero~\citep{parmar2023zero}, MagicBrush~\cite{magicbrush}, and Masactrl~\cite{cao2023masactrl}. The editing scenario here includes attribute editing, object replacement, style transfer and background editing.
Our ZZEdit paradigm can consistently improve the performance of P2P and PnP. Compared with other state-of-the-art methods, our ZZEdit shows its superiority through better background fidelity and editing consistency. More comparisons of editing results can be found in Appendix.

\section{Conclusion}
We presented a novel zero-shot image editing paradigm, dubbed ZZEdit. Given that intermediate-inverted latents can deliver a better trade-off between editability and fidelity than $\z_T$, we proposed to use a qualified $\z_p$ as editing pivot, which is sufficient-for-editing while structure-preserving. Then, a ZigZag process was designed to execute sampling and inversion alternately, which mildly approaches the target without ruining the structure information of $p$ step. Finally, we conducted a pure sampling process for the same inversion and sampling steps. Generally, our ZZEdit achieves better editing by fewer fidelity errors than the "inversion-then-editing" pipeline. Comprehensive experiments have shown that we achieve outstanding outcomes across a broad spectrum of text-driven image editing methods.





\normalem
\bibliography{example_paper}
\bibliographystyle{icml2025}

\clearpage
\appendix
\setcounter{figure}{0}
\setcounter{table}{0}
\renewcommand{\thetable}{A\arabic{table}}
\renewcommand {\thefigure} {A\arabic{figure}}

This Appendix includes 5 sections. 
Sec.~\ref{01} provides more visualization cross-attention maps of intermediate-inverted latents towards the target prompt $\mathcal{P}_{tgt}$. Sec.~\ref{02} gives more ablation study results of the proposed ZZEdit. Sec.~\ref{03} illustrates more qualitative results to compare our results with state-of-the-art image editing methods. 
Sec.~\ref{04} discusses the additional UNet operations for locating a better editing pivot $\z_p$ than $\z_T$. Sec.~\ref{05} introduces the limitations and future work of our ZZEdit.

\section{More Visualization of Cross-attention}
\label{01}
We display more cross-attention maps of intermediate-inverted latent $\z_t$ towards the target prompt $\mathcal{P}_{tgt}$, where $t \in [0.2T, 0.4T, 0.6T, 0.8T, T]$.  As shown in Fig.~\ref{fig:attention}, we give examples of  \textit{attribute editing, object replacement, style transfer and background editing}. It can be seen that intermediate-inverted latents can provide a considerable editability compared with the fully-inverted latent $\z_T$, thus achieving a better trade-off between \textit{editability} and \textit{fidelity} than the fully-inverted latent $\z_T$.

\begin{figure}[t]
    \centering
    \includegraphics[width=0.47\textwidth]{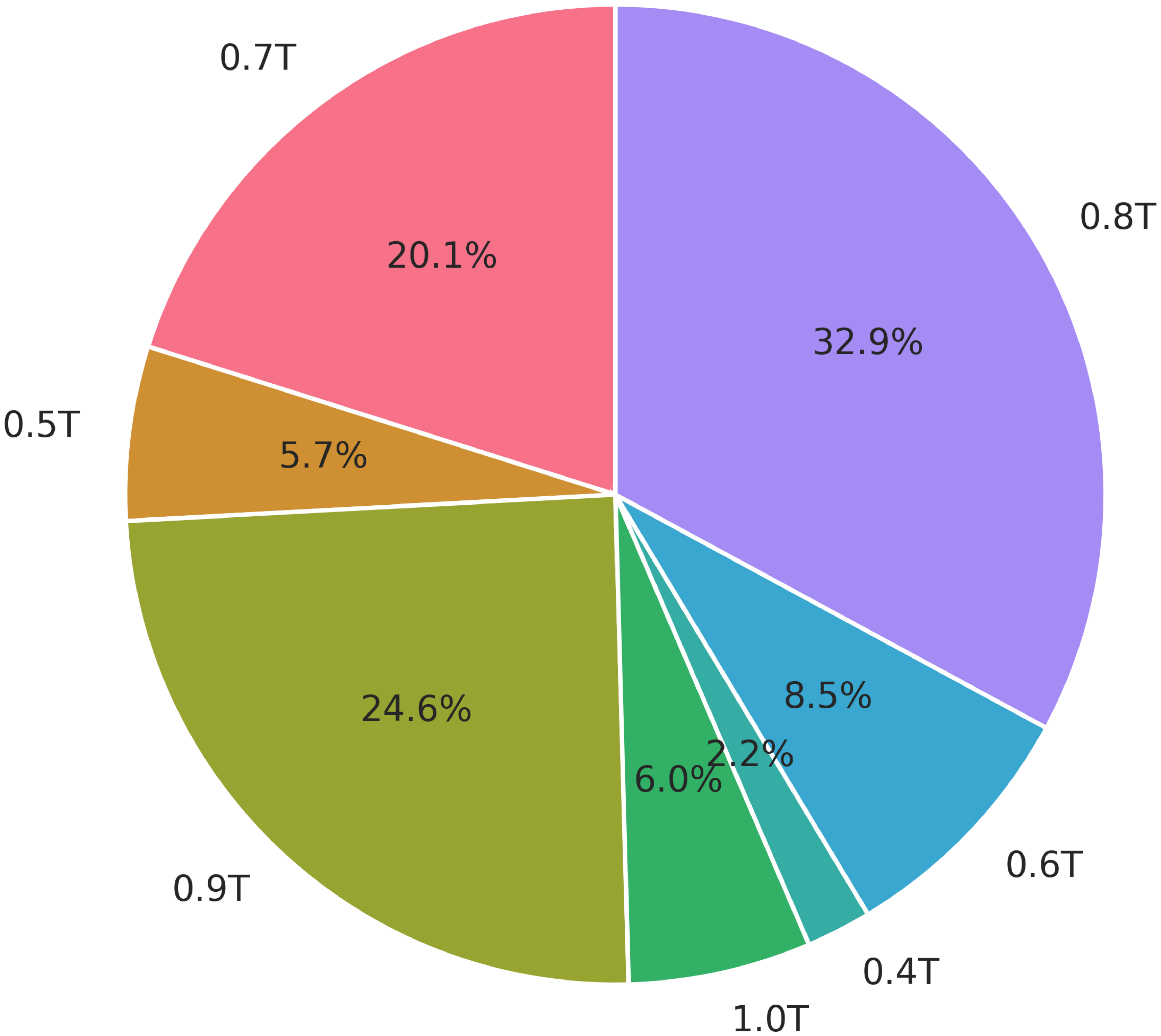}
    \caption{The statistics on the editing pivot $\z_p$ located by our ZZEdit on the  PIE-Bench dataset~\cite{ju2023direct}.}
    \label{fig:statistics}
\end{figure}
\begin{figure}[t]
    \centering
    \includegraphics[width=0.49\textwidth]{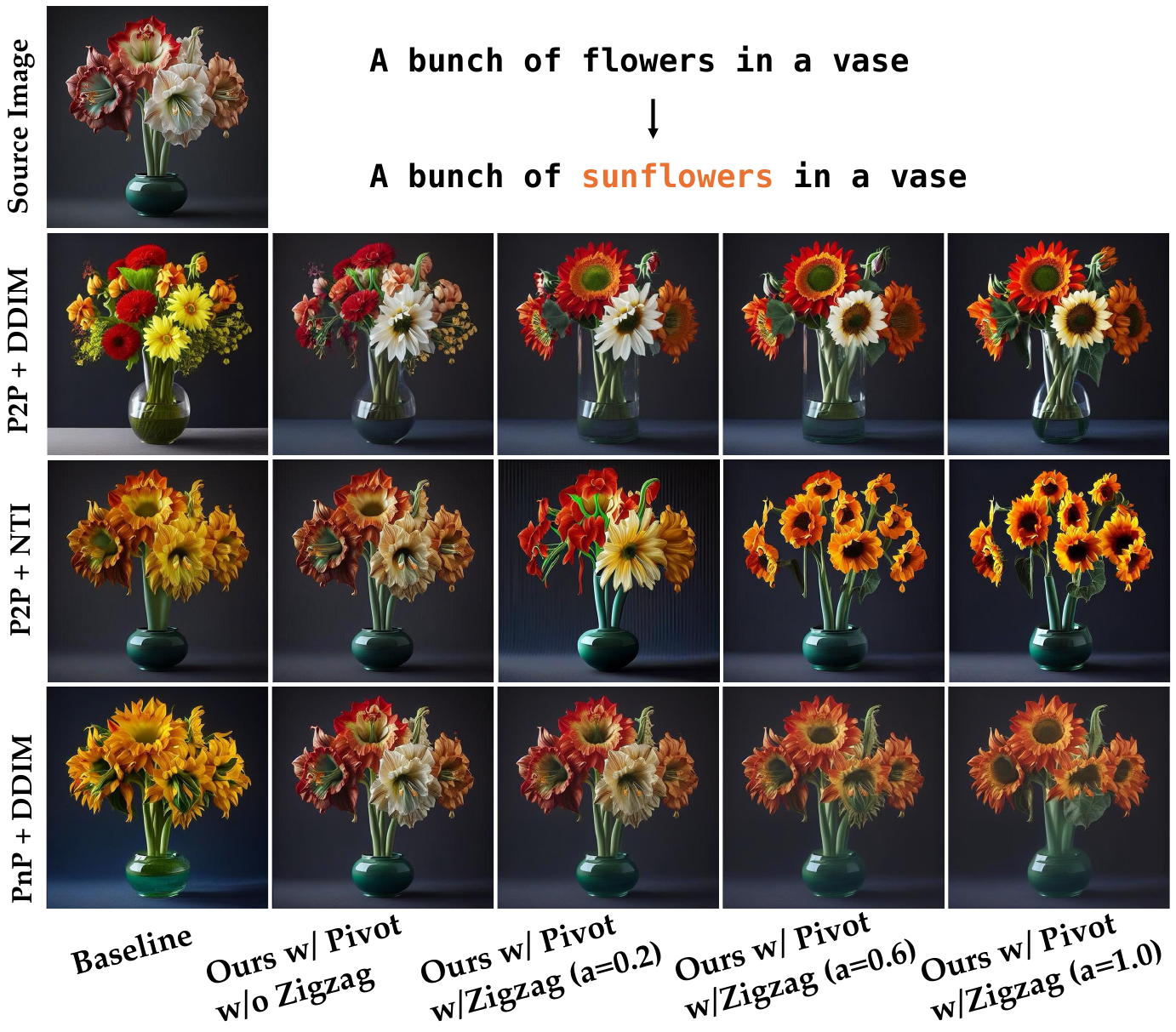}
    \caption{More qualitative ablation on our ZigZag process with P2P~\citep{p2p} and PnP~\citep{pnp}, which mildly enhances the guidance at a suitable pivot $\z_p$.
    }
    \label{fig:ablation2}
\end{figure}

\section{More Ablation Study}
\label{02}
\noindent{\textbf{Different Editing Pivots in ZZEdit.}}
We provide the visualization results using different points on the inversion trajectory as editing pivot in Fig. 4 of our main paper. Here, we display one more visualization example of editing the background from 'field' to 'beach' in Fig.~\ref{fig:sm_ab_diff}. We mark our located editing pivot ${\z}_p$ with purple.
Although the background corresponding to low-degree inversion is well maintained, its editability is insufficient. In contrast, a high-degree inversion brings editability but brings plentiful fidelity errors during reconstruction.
To better evaluate the effect of different editing pivots, 
as shown in Fig.~\ref{fig:gpt4-1} and Fig.~\ref{fig:gpt4-2},
we leverage GPT-4V(ision) system~\cite{gpt4-v}, which gives the editing comments by a Multimodal LLMs.

\begin{figure*}[t]
    \centering
    \includegraphics[width=1\textwidth]{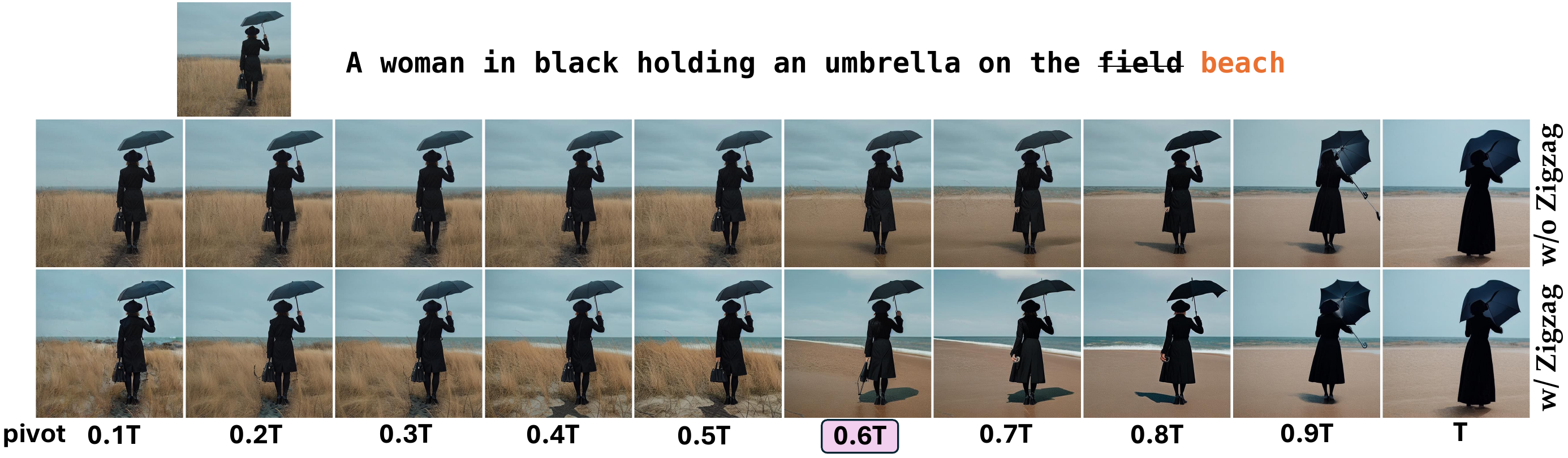}
    \caption{More ablation results of applying ZZEdit on P2P~\citep{p2p} \textit{w/} DDIM inversion, where different inverted latents are used with or without the ZigZag process equipped.}
    \label{fig:sm_ab_diff}
\end{figure*}


\noindent{\textbf{The Effectiveness and Distribution of Our Located Pivots.}}
In Tab. 1 of our main paper, we give the performance of selecting editing pivot from $[0.1T, 0.2T,...0.9T,T]$ randomly based on the P2P~\citep{p2p} \textit{w/} DDIM inversion, where a standard ZigZag process ($a=1$) is equipped.
In Tab.~\ref{sm_tab:ablation}, we also report the corresponding performance using P2P \textit{w/} NTI~\citep{null}
and PnP~\citep{pnp} \textit{w/} DDIM inversion.
Random pivot provides excellent background and structure preservation, but very poor editability with a standard ZigZag process. In contrast, our located pivot with a standard ZigZag process shows better editing consistency. This demonstrates the efficiency of our located pivot. Besides, as seen in Fig.~\ref{fig:statistics}, we provide the distribution of the editing pivots in our ZZEdit on the PIE-Bench dataset~\citep{ju2023direct}. Note that to save time and computation, we only look for the pivot from $[0.4T,0.5T,...0.9T,T]$ in practice. When the pivot reaches $T$ (i.e., $p=T$), our ZZEdit degenerates into the typical ``inversion-then-editing'' pipeline.

\begin{table*}[h]
\caption{Quantitative ablation study on the proposed ZigZag process with PnP~\citep{pnp} \textit{w/} DDIM inversion and P2P~\citep{p2p} \textit{w/} Null-text inversion. Results are obtained on the PIE-Bench dataset~\citep{ju2023direct}. We mark the best results of ZZEdit using located proper pivot $\z_p$ in bold. Here, the results of random pivot with the ZigZag process are also provided. Using random pivot shows poor editing consistency even though it has promissing background fidelity.
}
    \centering
    \small
\begin{tabular}{c|l|c|c|c|c|c|c|c}
  \toprule[1pt]

\multicolumn{2}{c|}{\multirow{2}{*}{\textbf {Method}}} &\textbf{Structure} & \multicolumn{4}{c|}{\textbf{Background Preservation}} &\multicolumn{2}{c}{\textbf{CLIP Similariy}}\\
\cline{3-9}

\multicolumn{2}{c|}{} & \textbf{L2} $\downarrow$  &\textbf{PSNR}$\uparrow$ &\textbf{LPIPS} $\downarrow$ &\textbf{MSE} $\downarrow$ &\textbf{SSIM} $\uparrow$  &\textbf{Whole}$\uparrow$ &\textbf{Edited}$\uparrow$  \\
\hline

\multicolumn{2}{c|}{\textbf{PnP+DDIM Baseline}}  &28.22  &22.28 &113.46 &83.64 &79.05 &25.41 &22.62  \\ 
\hline
\multirow{4}{*}{\bm{$w/$}  \textbf{Pivot}}
 &\bm{$w/o$} \textbf{ZigZag} (\bm{$a=0$})   &\textbf{19.37}  &\textbf{25.48} &\textbf{77.91} &\textbf{50.11} &
\textbf{83.09} &24.94 &22.22\\ 
 &\bm{$w/$} \textbf{ZigZag} (\bm{$a=0.2$})  &20.06  &25.29 &79.94 &50.99 &82.91 &25.00 &22.33\\ 
 &\bm{$w/$} \textbf{ZigZag} (\bm{$a=0.6$})    &21.94  &24.86 &84.69 &54.01 &82.41 &25.11 &22.54\\
 &\bm{$w/$} \textbf{ZigZag} (\bm{$a=1$})     &23.46  &24.55 &86.10 &55.04 &82.18 &\textbf{25.43} &\textbf{22.91}\\
 \hline
\multicolumn{2}{c|}{\textbf{Random Pivot} \bm{$w/$} \textbf{ZigZag} (\bm{$a=1$})}   &{12.53}  &{27.16} &{66.57} &{35.43} &{83.91} &24.16 &21.30 \\ 
\hline
\hline
\multicolumn{2}{c|}{\textbf{P2P+NTI Baseline}}  &13.44  &27.03 &60.67 &35.86 &84.11 &24.75 &21.86  \\ 
\hline
\multirow{4}{*}{\bm{$w/$}  \textbf{Pivot}}
 &\bm{$w/o$} \textbf{ZigZag} (\bm{$a=0$})  &\textbf{4.97}  &\textbf{29.79} &\textbf{36.62} &\textbf{19.89} &\textbf{86.71} &23.93 &20.94\\ 
 &\bm{$w/$} \textbf{ZigZag} (\bm{$a=0.2$})  &5.20 &29.64 &37.17 &20.14  &86.66 &23.99 &21.08\\ 
 &\bm{$w/$} \textbf{ZigZag} (\bm{$a=0.6$}) &{11.47} &{27.42} &{53.92} &{31.23} &{84.98} &24.95 &22.01 \\
 &\bm{$w/$} \textbf{ZigZag} (\bm{$a=1$}) &16.15  &26.67&84.28&49.06&82.14 &\textbf{25.16}&\textbf{22.13}\\
 \hline
\multicolumn{2}{c|}{\textbf{Random Pivot} \bm{$w/$} \textbf{ZigZag} (\bm{$a=1$})}  &14.72  &26.29 &76.71 &44.47 &82.72 &24.44 &21.43 \\    
\bottomrule[1pt] 
\end{tabular}
\label{sm_tab:ablation}
\end{table*}

\noindent{\textbf{The Effectiveness of The ZigZag Process.}}
As seen in Tab.~\ref{sm_tab:ablation}, we give the corresponding quantitative ablation results using PnP \textit{w/} DDIM inversion and P2P \textit{w/} NTI. With the increase of $a$, our proposed Zigzag process gradually increases editing consistency, thus obtaining better CLIP similarity. Besides, Fig.~\ref{fig:ablation2} shows a qualitative comparison on different baselines. Our ZZEdit can mildly approach the editing purpose through the increasing number of ZigZag steps ($a=0.2, 0.6$, and $1$) while still holding a satisfying background.

\begin{figure*}[t]
    \centering
    \includegraphics[width=0.9\textwidth]{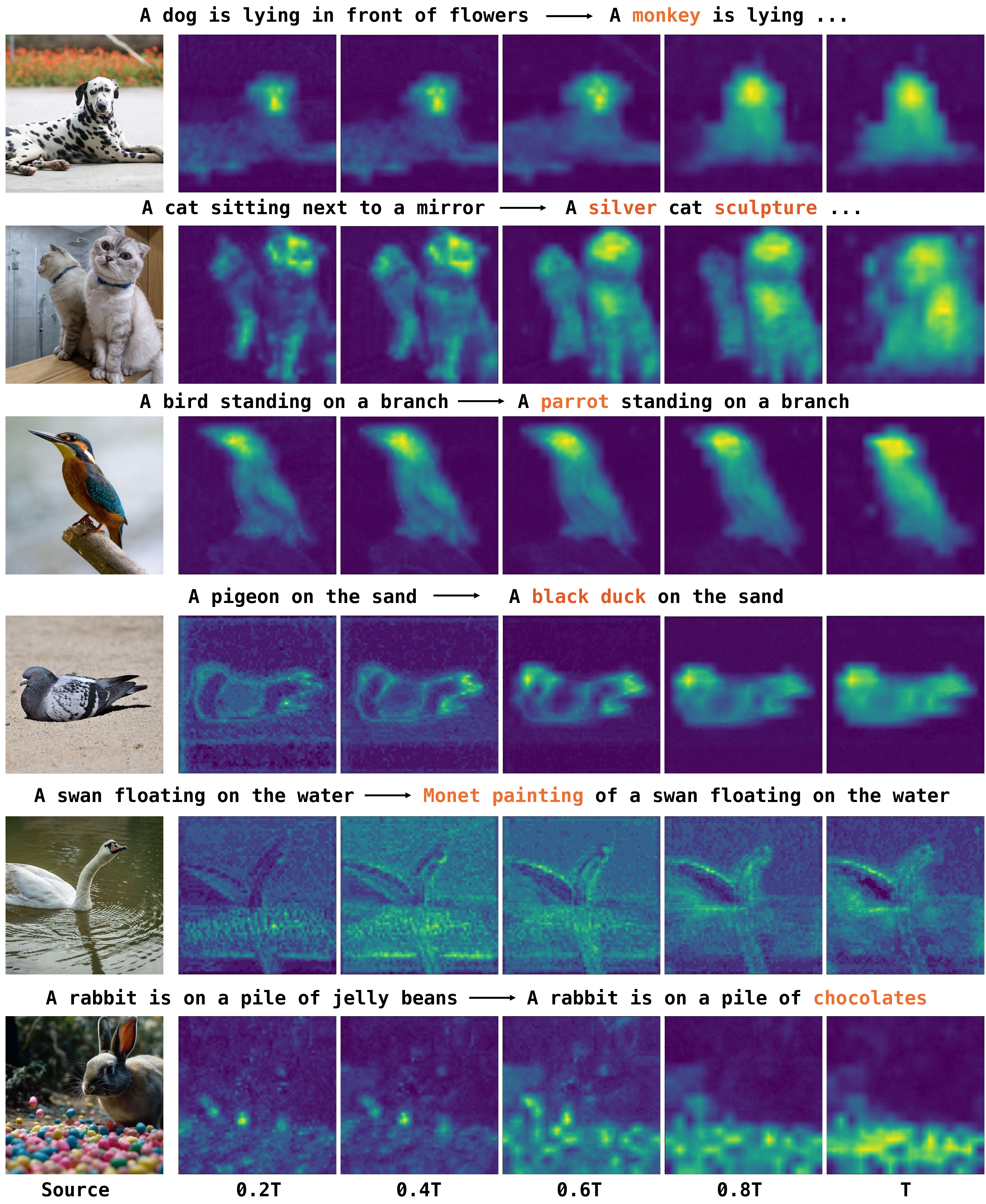}
    \caption{The cross-attention maps between different inverted latents $\z_t$ and the target prompt $\mathcal{P}_{tgt}$.}
    \label{fig:attention}
\end{figure*}

\section{More Image Editing Results}
\label{03}
As shown in Fig.~\ref{fig:sm_qualitative}, we show more qualitative comparison with the current text-driven editing methods, including P2P~\citep{p2p} \textit{w/} DDIM inversion and \textit{w/} NTI, PnP~\citep{pnp} \textit{w/} DDIM inversion, Pix2Pix-Zero~\citep{parmar2023zero}, MagicBrush~\citep{magicbrush}, and Masactrl~\citep{cao2023masactrl}.
The editing scenario here includes \textit{attribute editing, object replacement, style transfer and background editing}. Note that P2P \textit{w/} NTI often suffers from the color leak issue (see the 1st and 5th examples). The improvements are mostly tangible, and we circle some of the subtle discrepancies of the P2P and PnP baselines and the other compared methods in red.

\section{Additional UNet operations for Locating a Better Pivot}
\label{04}
Our ZZEdit paradigm needs to find a suitable editing pivot ${\z}_p$ before conducting ZigZag process for iterative manifold constraint, which takes additional UNet operations. Recall that we use Eqn.1 in our main paper for one-step DDIM sampling, obtaining the denoised latent $\hat{\z}_{t-1}$, $\bar{\z}_{t-1}$, and $\Tilde{\z}_{t-1}$ under the source prompt $\mathcal{P}_{src}$, null text $\varnothing$, and target prompt, respectively. Then, a qualified step $p$ is located by Eqn. 4. Here, in practice, we only look for the pivot from 7 options of $[0.4T,0.5T,...,0.9T,T]$. Thus, the maximum additional UNet operations are: $7*3=21$. Generally speaking, on a single Tesla A100 GPU, it takes about 23 seconds on average for an input image to seek such a qualified intermediate-inverted latent $\z_p$ as editing pivot.

\section{Limitations and Future Work}
\label{05}
While our method achieves promising results, it still faces some limitations. For example, we mainly apply ZZEdit into P2P and PnP, where the baseline model cannot generate new motion (e.g., ‘standing’ $\to$ ‘fly’). Our ZZEdit designs a dynamic latent trajectory for better editing performance, which cannot endow these baseline models with motion-editing capacities.

We find that GPT-4V~\citep{gpt4-v} can act as a good editing evaluator, so we hope to use it to build a new GPT-4V evaluation metric for text-driven image editing in the future.
Besides, for further motion editing, we will leverage our ZZEdit paradigm on the generic pretrained diffusion model for motion editing abilities.

\begin{figure*}[t]
    \centering
    \includegraphics[width=0.9\textwidth]{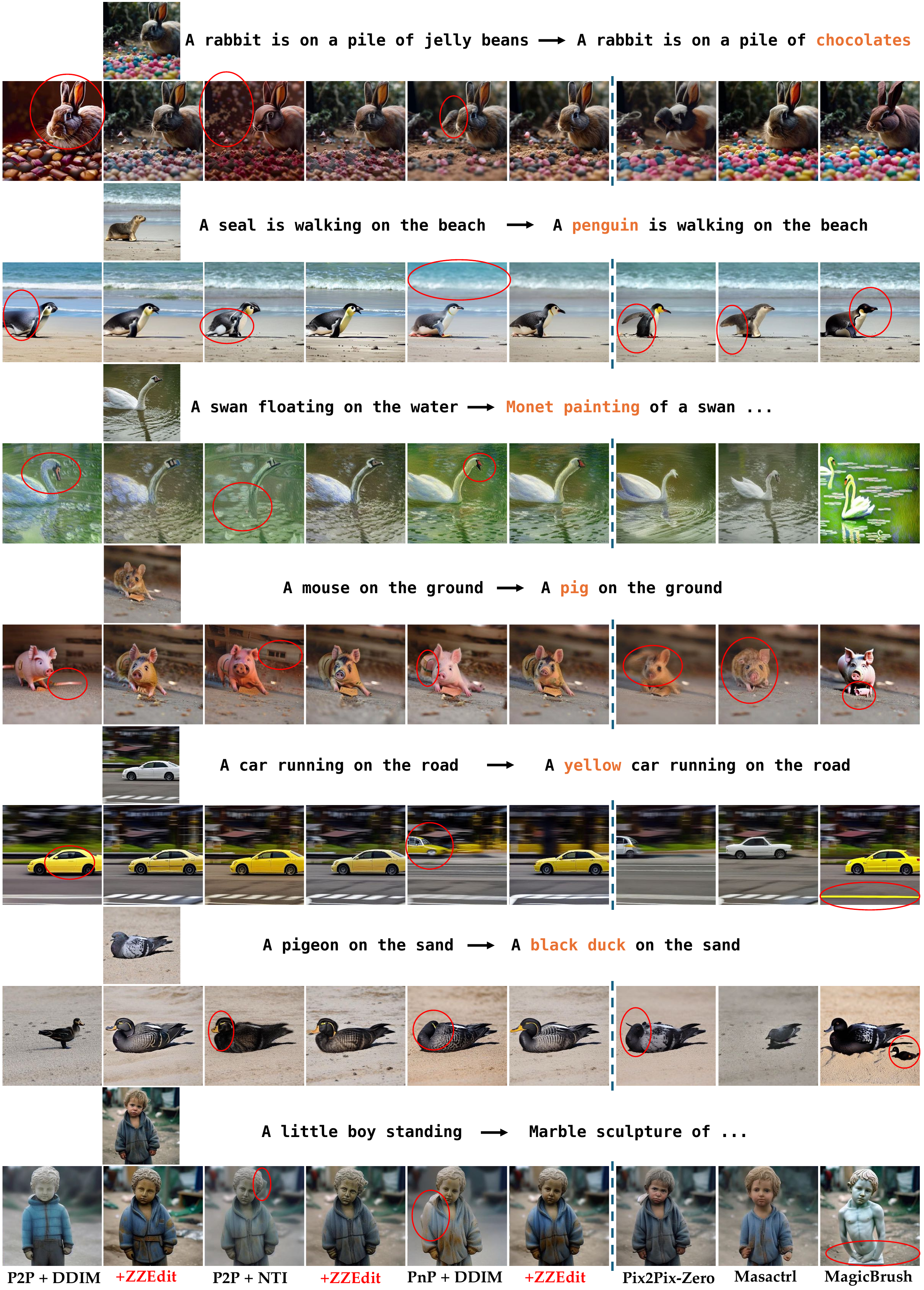}
    \caption{More visualization results of different editing techniques.}
    \label{fig:sm_qualitative}
\end{figure*}
\begin{figure*}[t]
    \centering
    \includegraphics[width=0.94\textwidth]{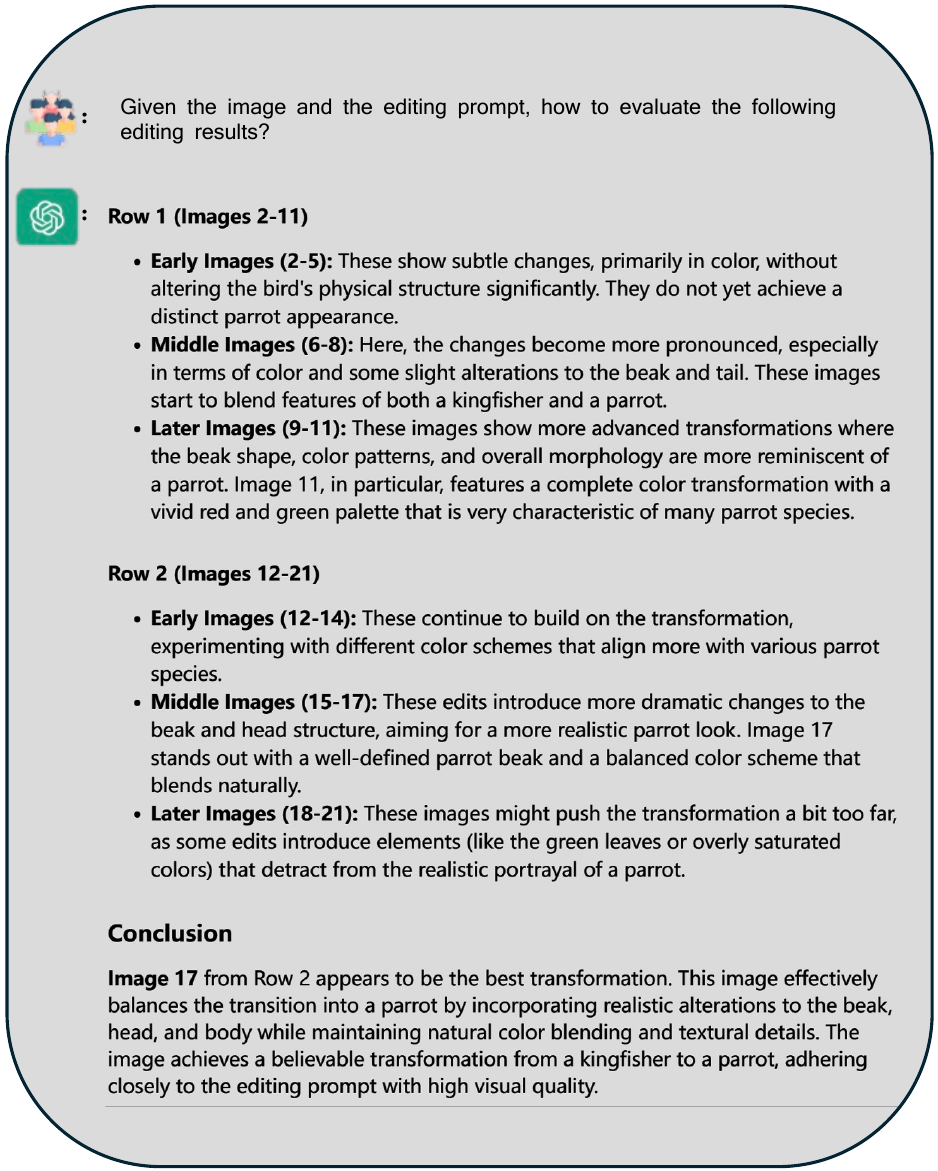}
    \caption{Using GPT-4V(ision) system~\citep{gpt4-v} for evaluating the editing example of Fig. 4 in our main paper. Here, we explore the effect of using different inversion-degree latent as the editing pivot with or without the ZigZag process equipped. We suggest using Fig. 4 as a reference.
    }
    \label{fig:gpt4-1}
    \vspace{-1em}
\end{figure*}
\begin{figure*}[t]
    \centering
    \includegraphics[width=0.94\textwidth]{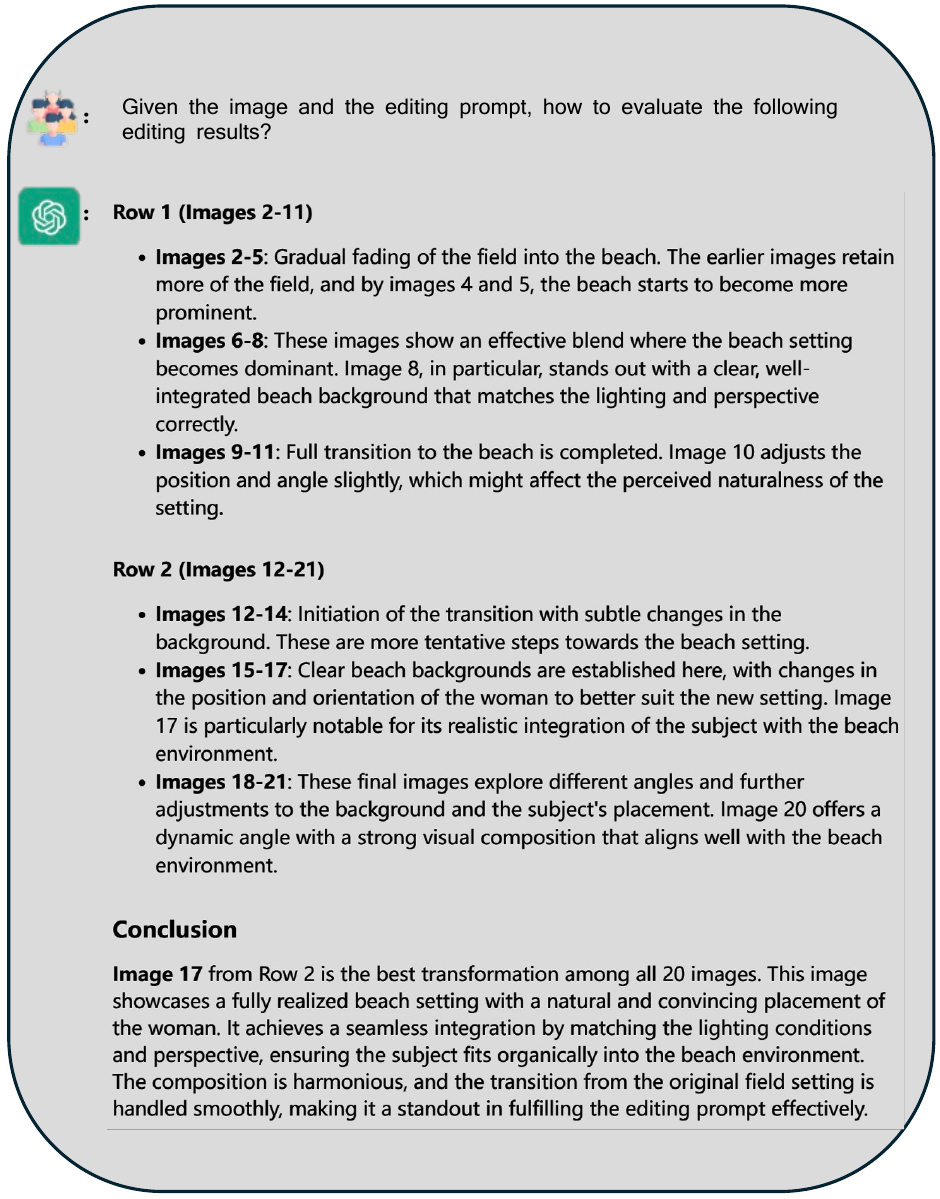}
    \caption{Using GPT-4V(ision) system~\citep{gpt4-v} for evaluating the editing example of Fig.~\ref{fig:sm_ab_diff} in this supplement. Here, we explore the effect of using different inversion-degree latent as the editing pivot with or without the ZigZag process equipped. We suggest using Fig.~\ref{fig:sm_ab_diff} as reference.
    }
    \label{fig:gpt4-2}
    \vspace{-1em}
\end{figure*}



\end{document}